\documentclass{article}

\PassOptionsToPackage{numbers, compress}{natbib}

\usepackage[final]{neurips_2023}

\usepackage[utf8]{inputenc} 
\usepackage[T1]{fontenc}    
\usepackage{hyperref}       
\usepackage{url}            
\usepackage{booktabs}       
\usepackage{amsfonts}       
\usepackage{nicefrac}       
\usepackage{microtype}      
\usepackage{xcolor}         

\usepackage{amsmath}
\usepackage[export]{adjustbox}
\usepackage{graphicx} 
\graphicspath{ {./figures/} }
\usepackage{wrapfig}
\usepackage{caption}
\usepackage{subcaption} 
\usepackage{bm}
\usepackage{multirow}
\usepackage{placeins}

\usepackage[sectionbib]{chapterbib}

\title{A Dual-Stream Neural Network Explains the Functional Segregation of Dorsal and Ventral Visual Pathways in Human Brains}

\usepackage{authblk}

\author[1]{Minkyu Choi}
\author[1]{Kuan Han}
\author[2]{\\Xiaokai Wang}
\author[1,3]{Yizhen Zhang}
\author[1,2]{Zhongming Liu}

\affil[1] {
  Department of Electrical Engineering and Computer Science\\
  University of Michigan\\
  Ann Arbor, MI 48109}
\affil[2] {
  Department of Biomedical Engineering\\
  University of Michigan\\
  Ann Arbor, MI 48109}
\affil[3] {
  Department of Neurological Surgery \\
  University of California, San Francisco\\
  San Francisco, CA 94143
}
\affil[ ]{\ttfamily\{cminkyu, kuanhan, xiaokaiw, zhyz, zmliu\}@umich.edu}

\begin{document}

\maketitle


\begin{abstract}
  The human visual system uses two parallel pathways for spatial processing and object recognition. In contrast, computer vision systems tend to use a single feedforward pathway, rendering them less robust, adaptive, or efficient than human vision. To bridge this gap, we developed a dual-stream vision model inspired by the human eyes and brain. At the input level, the model samples two complementary visual patterns to mimic how the human eyes use magnocellular and parvocellular retinal ganglion cells to separate retinal inputs to the brain. At the backend, the model processes the separate input patterns through two branches of convolutional neural networks (CNN) to mimic how the human brain uses the dorsal and ventral cortical pathways for parallel visual processing. The first branch (WhereCNN) samples a global view to learn spatial attention and control eye movements. The second branch (WhatCNN) samples a local view to represent the object around the fixation. Over time, the two branches interact recurrently to build a scene representation from moving fixations. We compared this model with the human brains processing the same movie and evaluated their functional alignment by linear transformation. The WhereCNN and WhatCNN branches were found to differentially match the dorsal and ventral pathways of the visual cortex, respectively, primarily due to their different learning objectives, rather than their distinctions in retinal sampling or sensitivity to attention-driven eye movements. These model-based results lead us to speculate that the distinct responses and representations of the ventral and dorsal streams are more influenced by their distinct goals in visual attention and object recognition than by their specific bias or selectivity in retinal inputs. This dual-stream model takes a further step in brain-inspired computer vision, enabling parallel neural networks to actively explore and understand the visual surroundings. 
\end{abstract}

\section{Introduction}

The human visual system comprises two parallel and segregated streams of neural networks: the "where" stream and the "what" stream \cite{mishkin1983object}. The "where" stream originates from magnocellular retinal ganglion cells and extends along the dorsal visual cortex. The "what" stream originates from parvocellular retinal ganglion cells and extends along the ventral visual cortex \cite{merigan1993parallel}. The two streams exhibit selective responses to different aspects of visual stimuli \cite{nassi2009parallel}. The "where" stream is tuned to coarse but fast information from a wide view, while the "what" stream is selective to fine but slow information from a narrow view \cite{merigan1993parallel, livingstone1988segregation}. The two streams are thought to serve different purposes. The "where" stream zooms out for spatial analysis \cite{haxby1991dissociation}, visual attention \cite{rizzolatti2003two, corbetta2002control, deco2004neurodynamical}, and guiding actions \cite{goodale1992separate} such as eye movements \cite{colby1999space}, while the "what" stream zooms in to recognize the object around the fixation \cite{tanaka1996inferotemporal}. While being largely parallel, the two streams interact with each other \cite{milner2017two}. In one way of their interaction, the "where" stream decides where to look next and guides the "what" stream to focus on a salient location for visual perception. As the eyes move around the visual environment, the interaction between the "where" and "what" streams builds a scene representation by accumulating object representations over time and space. This dual-stream architecture allows the brain to efficiently process visual information and support dynamic visual behaviors \cite{itti2000saliency}. 

\begin{figure}
  \centering
  \includegraphics[width=0.8\linewidth]{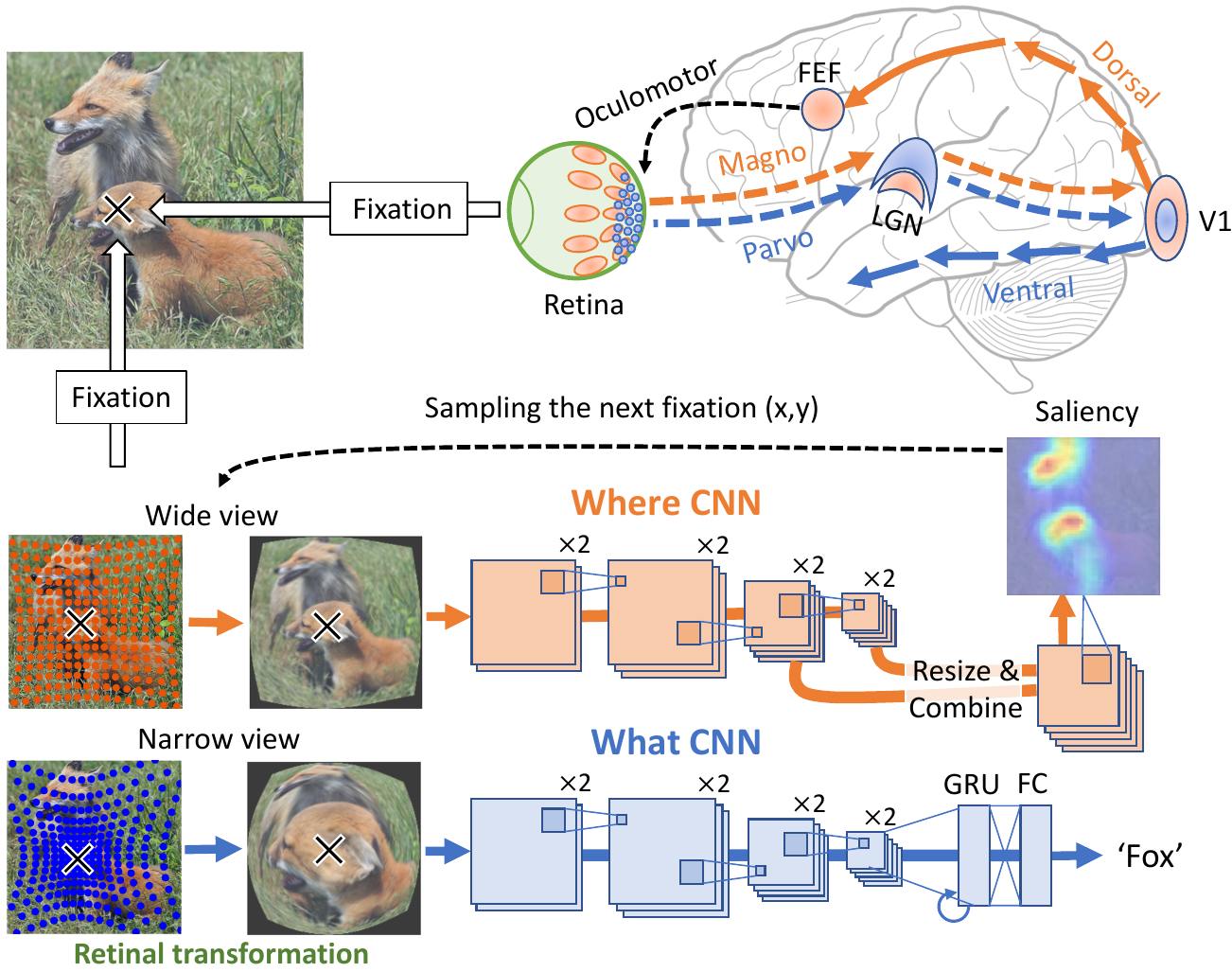}
  \caption{
   \textbf{Brain-inspired dual-stream vision model.} The top illustrates the subcortical (dashed arrows) and cortical (solid arrows) pathways for parallel visual processing in the brain. Given a scene (e.g., "two foxes on the lawn"), the retina samples incoming light relative to the fixation of the eyes (shown as the cross). Magnocellular (orange) and parvocellular (blue) retinal ganglion cells encode complementary visual information into two sets of retinal inputs relayed onto separate layers in the lateral geniculate nuclei (LGN) and further onto different neurons in the primary visual cortex (V1). Within V1, the relative ratio of magnocellular vs. parvocellular projections is higher for the periphery and lower for the fovea. Beyond V1, the magnocellular pathway continues along the dorsal visual cortex towards the intraparietal areas and further onto the frontal eye field (FEF) for oculomotor control, while the parvocellular pathway continues along the ventral visual cortex towards the inferior temporal cortex and further onto the superior temporal areas for semantic cognition. The bottom illustrates our model architecture including WhereCNN and WhatCNN. The model's frontend mimics the human retina and generates two separate input patterns relative to the fixation. One pattern is wider but coarser while the other is narrower but finer, providing the respective inputs to WhereCNN and WhatCNN. With the wide-view input, WhereCNN generates a probability map of saliency from which the next fixation is sampled. With a narrow-view input, WhatCNN generates an object representation per each fixation and constructs a scene representation recurrently from multiple fixations.
  }
  \label{fig:models}
\end{figure}

In contrast, computer vision systems tend to use a single stream of feedforward processing, acting as passive observers that sample visual information all at once with fixed and uniform patterns \cite{krizhevsky2017imagenet, he2016deep, dosovitskiy2020image}. Compared to human vision, this processing is less robust, especially given adversarial attacks \cite{szegedy2013intriguing, goodfellow2014explaining}; it is less efficient since it samples visual information equally regardless of salience or nuisance \cite{choi2022human}; it is less adaptive, lacking spatial attention for active sensing \cite{itti1998model, mnih2014recurrent}. These distinctions define a major gap between human and computer vision. Many visual tasks that are straightforward for humans are still challenging for machines \cite{lake2017building, davis2015commonsense}. Therefore, computer vision may benefit from taking further inspiration from the brain by using a dual-stream architecture to learn adaptive and robust visual behaviors. 

To gain insights into the computational mechanisms of human vision, researchers have developed image-computable models by utilizing goal-driven deep neural networks that simulate human perceptual behavior. In particular, convolutional neural networks (CNNs) are leading models of visual perception, capturing the hierarchical processing by the brain's ventral visual stream \cite{yamins2014performance, gucclu2015deep, eickenberg2017seeing, khaligh2014deep, yamins2016using}. Previous models of this nature commonly utilize CNNs trained through supervised learning \cite{yamins2014performance, khaligh2014deep, gucclu2015deep, cichy2016comparison, eickenberg2017seeing,  wen2018neural}, adversarial training \cite{dapello2020simulating, berrios2022joint}, unsupervised learning \cite{han2019variational, choi2018predictive}, or self-supervised learning \cite{zhuang2021unsupervised, wang2022incorporating,konkle2020deepnets}. However, models of the dorsal stream remain relatively under-explored, despite few studies \cite{rideaux2020but, mineault2021your, gucclu2017increasingly, bakhtiari2021functional}. Existing testing of these models has primarily focused on static images presented briefly to the fovea, thus limiting their assessment to a narrow range of visual behaviors and processes \cite{serre2019deep}. A more comprehensive approach is needed to develop models that incorporate both dorsal and ventral stream processing and to assess those models against brain responses when humans engage both the dorsal and ventral streams to freely explore complex and dynamic visual environments, which may be simulated in experimental settings \cite{kim2020naturalistic}.

To meet this need, we have developed a dual-stream model to mimic the parallel ventral and dorsal streams in the human brain \cite{mishkin1983object, merigan1993parallel, nassi2009parallel, goodale1992separate}. The model includes two branches of convolutional neural networks: WhereCNN and WhatCNN, which share the same architecture but receive distinct visual inputs and generate different outputs. WhereCNN samples a wide view to learn spatial attention and where to direct the subsequent gaze, while WhatCNN samples a narrow view to learn object representations. By taking multiple gazes at a given scene, the model sequentially samples the salient locations and progressively constructs a scene representation over both space and time. To evaluate this dual-stream model as a model of the human visual system, we have tested its ability to reproduce human gaze behavior and predict functional brain scans from humans watching a movie with unconstrained eye movements. Our hypothesis is that the model's WhereCNN and WhatCNN branches can effectively predict the brain responses along the brain's dorsal and ventral visual pathways, respectively. In addition, we have also conducted experiments to evaluate the underlying factors contributing to the functional segregation of the brain's dorsal and ventral visual streams. Of particular interest were the relative contributions of retinal sampling, spatial attention, and attention-guided eye movement in shaping the function of the dorsal stream and its interplay with the ventral stream during dynamic natural vision. 

\section{Related Works}

\subsection{Dorsal-stream vision}

Image-computable models of the brain's dorsal stream have been relatively limited compared to models of the ventral stream. Previous work has attempted to model the dorsal stream by training deep neural networks to detect motion \cite{rideaux2020but} or classify actions \cite{gucclu2017increasingly} using video inputs. However, these models do not fully capture the neuroscientific understanding that the dorsal stream is involved in locating objects and guiding actions, leading to its designation as the "where" or "how" visual pathway. More recent work by Mineault et al. focused on training a dorsal-stream model to emulate human head movements during visual exploration \cite{mineault2021your}. Additionally, Bakhtiari et al. utilized predictive learning to train parallel pathways and observed the ventral-like and dorsal-like representations as an emergent consequence of structural segregation \cite{bakhtiari2021functional}. However, no prior work has explored neural network models that emulate how the dorsal stream learns spatial attention and guides eye movements for visual navigation.

\subsection{Spatial attention and eye movement}

Prior research in the field of computer vision has attempted to train models to attend to and selectively focus on salient objects within a scene \cite{mnih2014recurrent, sermanet2014attention, fu2017look}, rather than processing the entire scene as a whole. This approach aligns with the brain's mechanism of spatial attention, where the dorsal stream acts as a global navigator, and the ventral stream functions as a local perceiver. In line with this mechanism, previous studies have employed dual-stream neural networks that process global and local features in parallel, aiming to achieve enhanced computational efficiency as a unified system \cite{esteves2017polar, sermanet2014attention, wang2020glance, guo2019global, wu2018learning}. However, these models do not fully replicate the way human eyes sample visual inputs during active exploration of the scene and thus still fall short in biological relevance.

\subsection{Foveated vision and retinal transformation}

The human retina functions as a sophisticated camera that intelligently samples and transmits visual information. It exhibits the highest visual acuity in the central region of the visual field, a phenomenon referred to as foveated vision \cite{derrington1984spatial,connolly1984representation,curcio1990human}. In contrast, peripheral vision uses lower spatial acuity but higher temporal sensitivity, making it better suited for detecting motion. These properties of retinal sampling are potentially useful for training neural networks to enhance performance \cite{min2022peripheral, thavamani2021fovea, jonnalagadda2021foveater, akbas2017object} or robustness \cite{harrington2021finding, vuyyuru2020biologically, choi2022human}, augment data or synthesize images \cite{wang2021use}. The retina also transmits information to the brain using distinct types of cells. The magnocellular and parvocellular retinal ganglion cells have different distributions, selectivity, and relay information through largely separate pathways. Taken together, the retina transforms visual information into segregated inputs for parallel visual processing. This biological mechanism has not been systematically investigated. 

Unlike the above prior works, our work combines multiple biologically inspired mechanisms into an integral learnable model. It uses a frontend inspired by the human retina and applies complementary retinal sampling and transformation. It uses two parallel pathways inspired by the human dorsal and ventral streams. It uses spatial attention and object recognition as the distinct learning objectives for training the two pathways. It further uses the attention to move fixations and thus allows the two pathways to interact for active sensing. Although these mechanisms have been explored separately in prior studies, their combination is novel and motivates our work to build and test such a model against the human brain and behavior in a naturalistic condition of freely watching a movie. 

\section{Methods}
In our model, WhereCNN and WhatCNN serve distinct functions in processing objects within a scene. WhereCNN identifies the spatial location of an object, determining "where" it is situated, while WhatCNN focuses on recognizing the identity of the object, determining "what" it is. When multiple objects are present in a scene, WhereCNN learns spatial attention and "how" to sequentially locate and fixate on each object. This allows the model to selectively attend to different objects in a sequence, mirroring the dynamic nature of human eye movements during visual exploration. Fig.\ref{fig:models} illustrates and describes the human visual system that inspires us to design our model.  

\subsection{Model design and training}
Akin to the human eyes \cite{brewer2002visual, gattass1981visual}, our model uses retinal transformation \cite{bashivan2019neural} to generate separate inputs to WhereCNN and WhatCNN. For both, the retinal input consists of $64\times 64$ samples non-uniformly distributed around the fixation. When describing a point in the retinal image and its corresponding point in the visual world in terms of the radial distance and the polar angle with respect to the fixation, their polar angles are the same while their radial distances are related by Eq.~\ref{eq:ret_trans}.

\begin{equation} \label{eq:ret_trans}
    r = g(r') = \frac{b}{\sqrt{\pi}}\frac{1 - \mathrm{exp}(\mathrm{ln}(a)r'/2)}{1 - \mathrm{exp}(\mathrm{ln}(a)/2)}
\end{equation}

where $r'$ and $r$ are the radial distances in the retinal and original images, respectively, $b$ is a constant that ensures $r_{\mathrm{max}}/g(r'_{\mathrm{max}})=1$, and $a$ controls the degree of center-concentration. Given a larger $a$, more retinal samples are closer to the fovea relatively to the periphery. We set $a=15$ for WhatCNN and $a=2.5$ for WhereCNN. In this setting, WhereCNN is more selective to global features, while WhatCNN is more selective to local features, mirroring the sampling bias of magnocellular and parvocellular retinal ganglion cells, as illustrated in Fig.\ref{fig:models}. 

Both WhereCNN and WhatCNN use similar backbone architectures. The backbone consists of four blocks of convolutional layers. Each block includes two \texttt{Conv2D} layers (kernel size $3\times 3$) followed by \texttt{ReLU} and \texttt{BatchNorm}. Applying $2\times 2$ \texttt{MaxPool} between adjacent blocks progressively reduces the spatial dimension. The feature dimension following each block is $64$, $128$, $256$, or $512$. Atop this backbone CNN, both WhereCNN and WhatCNN use additional components to support different goals. For WhereCNN, the feature maps from the 3rd and 4th convolutional blocks are resized to $16\times16$ and concatenated, providing the input to an additional convolutional block. Its output feature map is subject to \texttt{SoftMax} to generate a probability map of visual saliency. By random sampling by the probability of saliency, WhereCNN decides a single location for the next fixation. To avoid future fixations to revisit previously attended areas, inhibition of return (IOR) \cite{itti2001computational} is used. IOR keeps records of locations of prior visits as defined in Eq.\ref{eq:IOR}. 

\begin{equation}
    \textbf{ IOR}(t) = \textbf{ReLU} \Big( \bm{1} - \sum_{\tau=1}^{t} G(\bm{\mu}=\bm{l}_{\tau}, \bm{\Sigma} =\sigma^2\bm{I}) \Big)
    \label{eq:IOR}
\end{equation}

where $G(\bm{\mu},\bm{\Sigma})$ is a 2D Gaussian function centered at $\bm{l}_{\tau}$ (prior fixations) with a standard deviation $\sigma$ at the $\tau$-th step. Its values are normalized so that its maximum equals 1. By applying the IOR to the predicted saliency map using an element-wise multiplication, the future fixations would not revisit the areas already explored. 

For WhatCNN, the output feature map from the 4th convolutional block, after global average pooling, is given as the input to an additional layer of Gated Recurrent Units (GRU) \cite{chung2014empirical}, which recurrently update the representation from a sequence of fixations to construct a cumulative representation following another fully-connected layer. 

We first pre-train each stream separately and then fine-tune them together through three stages.

\textbf{Stage 1 - WhereCNN}. We first train WhereCNN for image recognition using ILSVRC2012 \cite{russakovsky2015imagenet} for object recognition and then fine-tune it for generating the saliency map to match human attention using SALICON dataset \cite{jiang2015salicon}. In this stage, we use random fixations to generate the retinal inputs to WhereCNN. Adam optimizer \cite{kingma2014adam} (lr=$0.002$, $\beta_1$=0.9, $\beta_2$=0.99) is used with $25$ epochs for SALICON training. At this stage, WhereCNN learns spatial attention from humans.

\textbf{Stage 2 - WhatCNN}. We first train WhatCNN for single-object recognition using ILSVRC2012 \cite{russakovsky2015imagenet} and then fine-tune it for multi-object recognition using MSCOCO \cite{lin2014microsoft}. In this stage, we use the pre-trained WhereCNN to generate a sequence of eight fixations and accordingly apply the retinal transformation to generate a sequence of retinal inputs to WhatCNN for recurrent object recognition. Note that the training in Stage 2 is confined to WhatCNN, while leaving WhereCNN as pre-trained in Stage 1. Adam optimizer (lr=$0.002$, $\beta_1$=0.9, $\beta_2$=0.99) is used with $40$ epochs for MSCOCO training. 

\textbf{Stage 3 - WhereCNN \& WhatCNN}. Lastly, we equally combine the two learning objectives to train both WhereCNN and WhatCNN altogether and end-to-end using eight fixations. Adam optimizer  (lr=$0.0002$, $\beta_1$=0.9, $\beta_2$=0.99) is used with $25$ epochs for training. In this stage, SALICON, which contain labels for both saliency prediction and object recognition, is used for training. More details can be found at Appendix B \footnote{The code and appendix are available at \url{https://github.com/minkyu-choi04/DualStreamBrains/}}.

\subsection{Model evaluation with human gaze behavior and fMRI responses} \label{sec_encoding}
We use two criteria to evaluate how well a model matches the brain given naturalistic and dynamic visual stimuli. First, the model should generate similar human visual behaviors, such as visual perception and gaze behavior. Second, the model's internal responses to the stimuli should predict the brain's responses to the same stimuli through linear projection implemented as linear encoding models \cite{naselaris2011encoding}. For our dual-stream model, we hypothesize that WhereCNN better predicts dorsal-stream voxels and WhatCNN better predicts ventral-stream voxels. 

For this purpose, we use a publicly available fMRI dataset from a prior study \cite{haxby2011common}, in which a total of 11 human subjects ($4$ females) were instructed to watch the movie Raiders of the Lost Ark ($115$ minutes) with unconstrained eye movements. The movie was displayed on an LCD projector with a visual angle of $17^\circ \times 22.7^\circ$. Whole-brain fMRI data was acquired in a 3-T MRI system with a gradient-recalled echo planar imaging sequence  (TR/TE = $2.5\mathrm{s}$/$35\mathrm{ms}$, flip angle = $90^\circ$, nominal resolution = $3\mathrm{mm}\times3\mathrm{mm}\times3\mathrm{mm}$). We preprocess the data by using the minimal preprocessing pipeline released by the Human Connectome Project (HCP) \cite{glasser2013minimal}

We test how well the model can predict the voxel-wise fMRI response to the movie stimuli through a learnable linear projection of artificial units in the model. To evaluate whether and how the two branches in the model differentially predict the two streams in the brain, we define two encoding models for each voxel: one based on WhereCNN and the other based on WhatCNN. We train and test the encoding models with data during different segments of the movie. To avoid overfitting, we apply dimension reduction to the internal responses in either WhereCNN or WhatCNN by applying principal component analysis (PCA) first to each layer and then to all layers while retaining $99$\% of the variance \cite{wen2018neural, han2019variational}. We further convolve the resulting principal components with a canonical hemodynamic response function (HRF) that peaks at $5$ seconds, down-sample them to match the sampling rate of fMRI, generating the linear regressors used in the encoding model. Using the training data ($81$\% of the total data), we estimate the encoding parameters using L2-regularized least squares estimation. Using the held-out testing data ($19$\%), we test the encoding models for their ability predicting the fMRI responses observed at each voxel and measure the accuracy of prediction as the correlation between the predicted and measured fMRI responses, denoted as $r_{where}$ and $r_{what}$ for the encoding models based on WhereCNN and WhatCNN. We test the significance of the prediction using a block permutation test \cite{adolf2014increasing} with a block size of $20$-seconds and $100,000$ permutations and apply the false discovery rate (FDR) ($p<0.05$). We further differentiate the relative roles of the brain's WhereCNN vs. WhatCNN in predicting the brain's dorsal and ventral streams for single voxels as well as regions of interest. For this, we define a relative performance (Eq.\ref{eq:pwhere}).

\begin{equation}
    p_{where} = \frac{r^2_{where}}{r^2_{where} + r^2_{what}}
    \label{eq:pwhere}
\end{equation}

In the range from $0$ to $1$, $p_{where}>0.5$ indicates better predictive performance by WhereCNN, while $p_{where}<0.5$ indicates better predictive performance by WhatCNN.


\subsection{Alternative models and control experiments}
By design, the WhereCNN and WhatCNN branches within our model exhibit two key distinctions. WhereCNN is specifically trained to learn spatial attention by utilizing wider views, while WhatCNN focuses on object recognition through the use of local views. To explore the impact of input views and learning objectives on the model's capacity to predict brain responses, we introduce two modified control streams: ControlCNN-a and ControlCNN-b, designed as hybrid variants encapsulating diverse input views and learning objectives. 
ControlCNN-a receives a narrower view as its input and is trained for saliency prediction. Conversely, ControlCNN-b is curated to accommodate a broader view and primarily focuses its training on object recognition. These adjustments yield a versatile examination of their respective influences and functionalities.
By combining WhereCNN or WhatCNN with ControlCNN-a or ControlCNN-b, we create four alternative dual-stream models (illustrated in Fig.\ref{fig:ablation_results_controlCNN}) and examine their abilities to explain the functional segregation of the brain's dorsal and ventral streams.

\section{Results}

\subsection{WhereCNN learns attention and WhatCNN learns perception}

\begin{figure}
  \centering
  \includegraphics[width=0.99\linewidth]{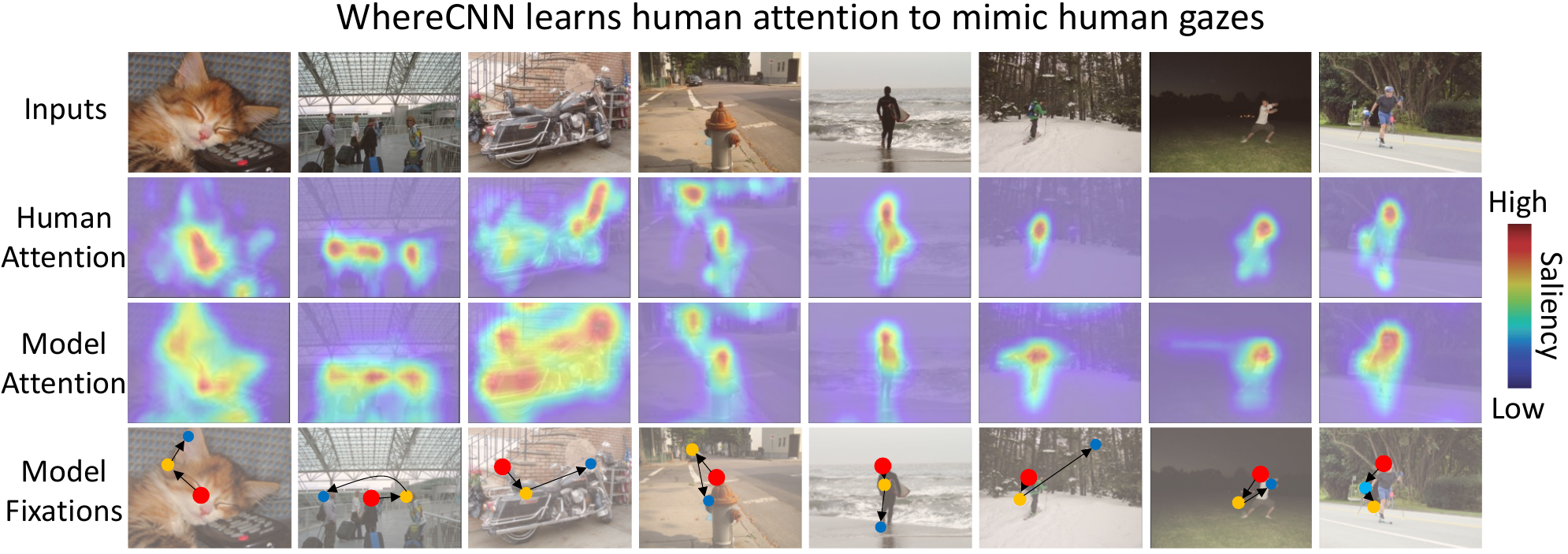}
  \caption{
   \textbf{Saliency prediction}. Given an image (1st row), WhereCNN generates a saliency map (3rd row) similar to the map of human attention (2nd row). Sampling this saliency map generates a sequence of fixations (as red/orange/blue circles in the order of time in the 4th row) similar to human saccadic eye movements (not shown).  
  }
  \label{fig:behavior_model}
\bigskip
\bigskip
\bigskip

  \centering
  \includegraphics[width=0.99\linewidth]{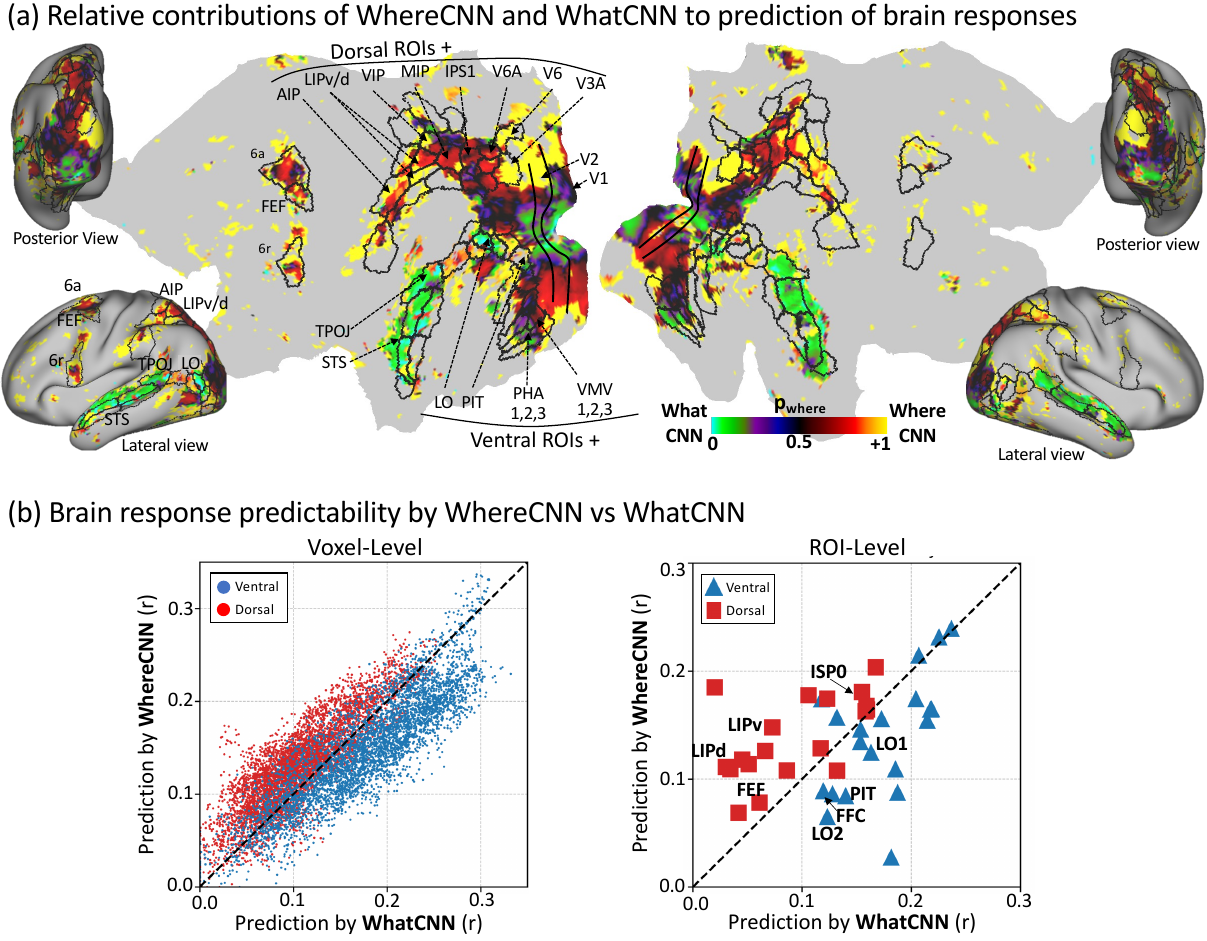}
  \caption{
   \textbf{Differential encoding of the dorsal and ventral streams}. (a) Relative contributions of WhereCNN and WhatCNN to the prediction of the fMRI response observed at each cortical location. Color highlights the locations significantly predictable by the model (FDR<0.05, block permutation test). The color itself indicates the degree by which WhereCNN is more predictive than WhatCNN (warm-tone) or the opposite (cool-tone). Visual areas are delineated and labeled based on brain altas \cite{glasser2016multi}. Panel (b) plots the predictive performance by WhereCNN (y-axis) against that by WhatCNN (x-axis) and shows a clear separation of voxels (left panel) or ROIs (right panel) along the dorsal stream (red) vs. ventral stream (blue) relative to the dashed line of equal predictability. See Appendix A for the full ROI labels. 
  }
  \label{fig:main_results_VD}
\end{figure}

The WhereCNN and WhatCNN branches in our model are specifically designed to fulfill different objectives: predicting human visual saliency and recognizing visual objects, respectively. In Fig.\ref{fig:behavior_model}, we present examples comparing human attention with the model's attention based on the SALICON's validation set. WhereCNN can successfully identify salient locations where humans are more likely to direct gaze. Additionally, WhereCNN can mimic human saccadic eye movements by generating a sequence of fixations that navigate the model's attention to those salient locations. In contrast, WhatCNN can recognize either single or multiple objects (macro F1 score on MSCOCO's validation set: 61.0).

\subsection{WhereCNN and WhatCNN matches dorsal and ventral visual streams} \label{sec:enc_main}
By using linear encoding models, we use the WhereCNN and WhatCNN branches to predict fMRI responses during the processing of identical movie stimuli by both the model and the brain. Together, these two branches can predict responses across a wide range of cortical locations involved in visual processing. However, they exhibit distinct predictive power in relation to the dorsal and ventral streams. Generally, the WhereCNN branch exhibits superior predictive performance for the dorsal stream, while the WhatCNN branch performs better in predicting responses within the ventral stream (Fig.\ref{fig:main_results_VD}). For the early visual areas (V1, V2, V3), WhereCNN better predicts the peripheral representations, while WhatCNN better predicts the foveal representations.

\begin{figure}
  \centering
  \includegraphics[width=0.999\linewidth]{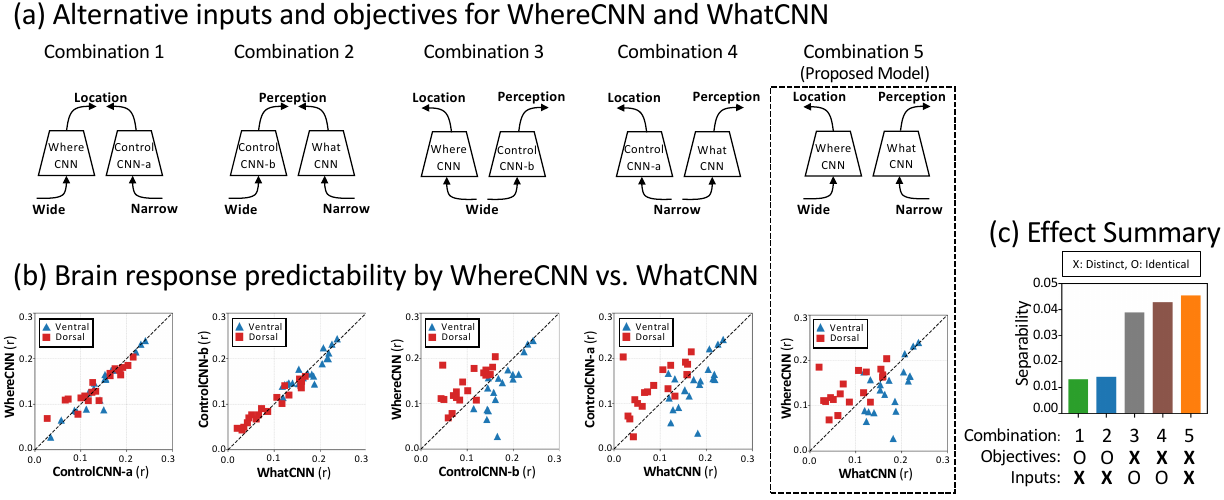}
  \caption{
   \textbf{Contributing factors of the dorsal-ventral functional segmentation.} 
   (a) Alternative designs of the two-stream model for investigating the contributing factors of the functional segregation of two streams in the proposed model. In addition to WhereCNN and WhatCNN in the proposed model, we included ControlCNN-a (narrow input field of view for predicting location) and ControlCNN-b (wide input field of view for predicting perception) for an ablation study.
   (b) The predictive performances and functional segregations by the two streams are plotted for the dorsal (red squares) vs. ventral (blue triangles) ROIs for each of the alternative models in (a), correspondingly. The dashed diagonal lines represent equal predictive abilities of ventral and dorsal ROIs.
   (c) Quantitative evaluation of the functional segregation of the dorsal and ventral ROIs relative to the dashed line of equal predictability. Separability measures how far the predictions are away from the dashed line. Assuming the coordinates of a certain ROI is ($x,y$), separability is calculated as the average of $|x-y|$ for all ROIs. 
   }
  \label{fig:ablation_results_controlCNN}
\end{figure}

\subsection{Factors underlying the functional segregation of the dorsal and ventral streams}
We further investigate the underlying factors contributing to the model's ability to explain the functional segregation of the brain's dorsal and ventral visual streams (as depicted in Fig.\ref{fig:main_results_VD}). Specifically, we examine the input sampling pattern and output learning objective, both of which are distinct for the WhereCNN and WhatCNN branches in our dual-stream model. 

To investigate the contributing factors of the functional segregation in predicting human ventral and dorsal streams, we introduce four variations of the proposed model, where the two branches either share their inputs or have the same learning objectives, and compare their abilities to account for the functional segregation of the dorsal and ventral visual streams (as shown in Fig. \ref{fig:ablation_results_controlCNN}). 
When the two branches solely differ in their input sampling, they are unable to explain the dorsal-ventral segregation (combination 1 and 2). However, when the two branches exclusively differ in their learning objectives, the functional segregation is better explained (combination 3 and 4). Moreover, when the two branches differ in both input sampling and learning objectives (combination 5), as utilized in our proposed model, the functional segregation is even more pronounced. These ablation experiments suggest that the distinct learning objectives of the brain's dorsal and ventral streams is the primary factor underlying their functional segregation. 

\subsection{Dual-stream: a better brain model than single-stream}
\begin{figure}
  \centering
  \includegraphics[width=0.99\linewidth]{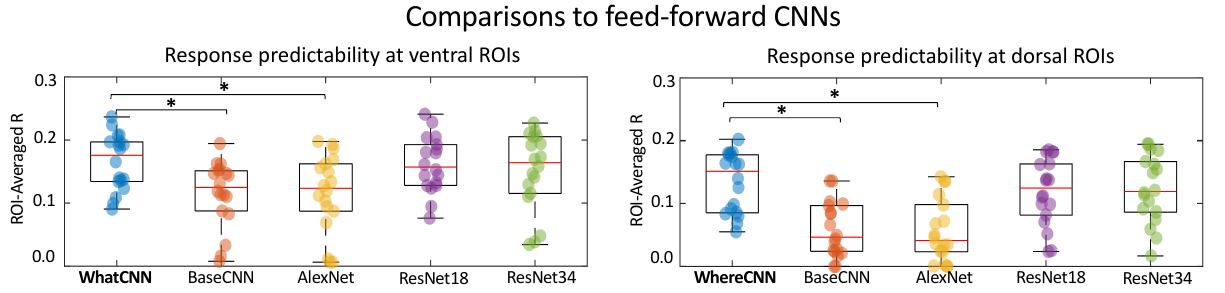}
  \caption{
   WhatCNN (left) or WhereCNN (right) vs. alternative single-stream CNNs. The boxplot shows the encoding performance of different models for ventral or dorsal visual areas. Each dot within the box plot signifies the average prediction accuracy r within a respective ROI in the ventral or dorsal region. Asterisk (*) represents a significant difference by the Wilcoxon signed-rank test ($\alpha=0.05$).
  }
  \label{fig:main_results_VD_bar}
\end{figure}

We also compare our dual-stream model with single-stream alternatives. One of these alternatives is a baseline CNN that shares the same backbone architecture as a single branch in our dual-stream model. However, this baseline CNN is trained with original (224x224) images to recognize objects in ImageNet \cite{russakovsky2015imagenet} and MS-COCO \cite{lin2014microsoft}. Thus, it serves as a direct comparison with either the WhereCNN or WhatCNN branch in our model. In addition, we also include AlexNet \cite{krizhevsky2017imagenet}, ResNet18, and ResNet34 \cite{he2016deep} as additional alternatives, which have been previously evaluated in relation to brain responses \cite{wen2018neural, wen2018deep}. We compare these single-stream alternatives with either branch in our model in terms of their ability to predict brain responses within dorsal or ventral visual areas (Fig.\ref{fig:main_results_VD_bar}). Despite their use of the same backbone architecture, the baseline under-performs WhatCNN in predicting responses in ventral visual areas, and under-performs WhereCNN in predicting responses in dorsal visual areas. This result suggests that the interactive and parallel nature of the dual-stream model renders each stream more akin to the functioning of the human brain, surpassing the performance of isolating a single stream. Moreover, WhatCNN or WhereCNN also performs better than AlexNet and comparably with ResNet18 and ResNet34, which are deeper than the architecture of our model.

\begin{figure}
  \centering
  \includegraphics[width=0.99\linewidth]{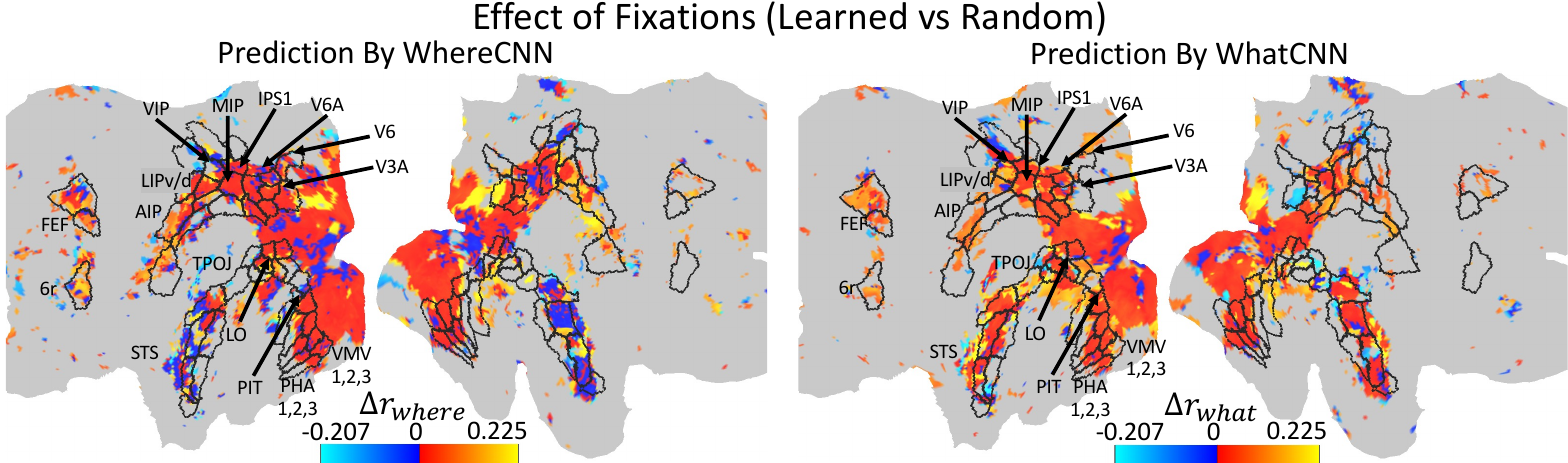}
  \caption{
   \textbf{Effects of attention-driven eye movements.} The use of attention to determine fixations vs. the use of random fixations is evaluated in terms of the resulting difference in the encoding performance by WhereCNN (left) and WhatCNN (right), denoted and color-coded as $\Delta r_{where}$ and $\Delta r_{what}$, respectively. Voxels displayed in warm colors indicate that the predictions are more accurate with learned fixations, whereas those in cool colors signify better predictions with random fixations.
  }
  \label{fig:main_results_fixs}
\end{figure}

\subsection{Attention-driven eye movements improve encoding}

Similar to human gaze behavior towards salient objects, our model learns spatial attention to guide fixations for parallel visual processing. In this study, we investigate whether and how the model's ability to predict brain responses depends on its utilization of attention-driven fixations. To examine this, we conduct experiments where the model is allowed to use either attention-driven fixations or random fixations to collect retinal samples, and we evaluate how this choice impacts the model's capability to predict brain responses by calculating $\Delta r = r^{learned} - r^{random}$ for all voxels, where $r^{learned}$ and $r^{random}$ are encoding performances of each voxel using learned or random fixations respectively. As depicted in Fig.\ref{fig:main_results_fixs}, employing attention-driven fixations leads to higher encoding accuracy by both WhereCNN and WhatCNN compared to the use of random fixations for a majority of visual cortical locations within both the dorsal and ventral streams.

\section{Discussion}
In summary, we introduce a new dual-stream neural network that incorporates the brain's mechanisms for parallel visual processing. The defining features of our model include 1) using retinal transformation to separate complementary inputs to each stream, 2) using different learning objectives to train each stream to learn either spatial attention or object recognition, and 3) controlling sequential fixations for active and interactive visual sensing and processing. We demonstrate that the combination of these features renders the model more akin to the human brain and better predictive of brain responses in humans freely engaged in naturalistic visual environments. Importantly, the two streams in our model differentially explain the two streams in the brain, contributing to the computational understanding as to how and why the brain exhibits and organizes distinct responses and processes along the structurally segregated dorsal and ventral visual pathways. Our findings suggest that the primary factor contributing to the dorsal-ventral functional segregation is the different goals of the dorsal and ventral pathways. That is, the dorsal pathway learns spatial attention to control eye movements \cite{bisley2010attention, corbetta2002control, yantis2003cortical, maunsell2006feature}, while the ventral stream learns object recognition. 

Although our model demonstrates initial steps to model parallel visual processing in the brain, it has limitations that remain to be addressed in future studies. For one limitation, the model uses different spatial sampling to generate the retinal inputs to the two streams but does not consider different temporal sampling that makes the dorsal stream more sensitive to motion than the ventral stream \cite{rideaux2020but, gucclu2017increasingly, mineault2021your, bakhtiari2021functional}. For another limitation, the interaction between the two streams is limited to the common fixation that determines the complementary retinal input to each stream. Although attention-driven eye movement is an important aspect of human visual behavior shaping brain responses for both dorsal and ventral streams, the two streams also interact and exchange information at higher levels. The precise mechanisms for dorsal-ventral interactions remain unclear but may be important to understanding human vision or improving brain-inspired computer vision.  


\section{Acknowledgements}

This research is supported by the Collaborative Research in Computational Neuroscience (CRCNS) program from National Science Foundation (Award\#: IIS 2112773) and the University of Michigan.

\end{document}